\documentclass[conference]{IEEEtran}
\IEEEoverridecommandlockouts

\newcommand{\paperTitle}{Robust, Extensible, and Fast: Teamed Classifiers for Vehicle Tracking and Vehicle Re-ID in Multi-Camera Networks}
\newcommand{\paperKeywords}{teamed classifiers, video analytics, vehicle re-id, conditional computation, vehicle re-identification, convolutional attention, targeted attention}
\newcommand{\paperAuthors}{Abhijit Suprem, Rodrigo Lima, Bruno Padilha, Joao Ferreira, Calton Pu}
\usepackage[hyphens]{url}
\usepackage{xstring}

\usepackage[usenames,dvipsnames]{xcolor}
\definecolor{linkcolor}{HTML}{647382}
\definecolor{citecolor}{HTML}{647382} %
\definecolor{urlcolor}{rgb}{0.4,0.2,0.2}
\definecolor{sqlcolor}{HTML}{965d67}
\definecolor{smtcolor}{HTML}{5d968c}

\usepackage[breaklinks]{hyperref}
\hypersetup{%
	pdfauthor = {\paperAuthors},
	pdftitle = {\paperTitle},
	pdfkeywords = {\paperKeywords},
	bookmarksopen = {true},
	colorlinks=true,
	citecolor={urlcolor},
	linkcolor={linkcolor},
	urlcolor={citecolor},
	pdfborder={ 0 0 0 }
}

\usepackage[usenames,dvipsnames]{xcolor}
\usepackage{amsmath,amsopn,amssymb}
\usepackage{listings}
\usepackage[caption=false]{subfig}
\usepackage{endnotes,microtype,xspace,graphicx,fancyvrb,multirow}
\usepackage{supertabular,booktabs}
\usepackage{array,underscore,relsize}
\usepackage[T1]{fontenc}
\usepackage{times}
\usepackage{fancyhdr,lastpage}
\usepackage{enumitem}
\usepackage{balance}
\usepackage{booktabs}
\usepackage{pifont}
\usepackage{listings}
\usepackage{multirow}
\usepackage[scaled]{beramono}
\usepackage{tabularx}
\usepackage{wrapfig}
\usepackage{lipsum}

\usepackage{stmaryrd}
\usepackage{ltablex}
\usepackage{mathtools}
\usepackage{mathrsfs}
\usepackage{ dsfont }

\usepackage{amsmath}
\usepackage{algpseudocode}
\usepackage[ruled,linesnumbered, vlined]{algorithm2e}
\usepackage{enumitem}
\usepackage{algpseudocode}

\usepackage{multirow}
\usepackage[normalem]{ulem}
\useunder{\uline}{\ul}{}

\usepackage[capitalize,noabbrev]{cleveref}

\def\BibTeX{{\rm B\kern-.05em{\sc i\kern-.025em b}\kern-.08em
    T\kern-.1667em\lower.7ex\hbox{E}\kern-.125emX}}

\newcommand{\squishitemize}{
 \begin{list}{$\bullet$}
  { \setlength{\itemsep}{0pt}
     \setlength{\parsep}{3pt}
     \setlength{\topsep}{3pt}
     \setlength{\partopsep}{0pt}
     \setlength{\leftmargin}{1.95em}
     \setlength{\labelwidth}{1.5em}
     \setlength{\labelsep}{0.5em} } }

\newcounter{Lcount}
\newcommand{\squishlist}{
    \begin{list}{\arabic{Lcount}. }
   { \usecounter{Lcount}
        \setlength{\itemsep}{0pt}
        \setlength{\parsep}{3pt}
        \setlength{\topsep}{3pt}
        \setlength{\partopsep}{0pt}
        \setlength{\leftmargin}{2em}
        \setlength{\labelwidth}{1.5em}
        \setlength{\labelsep}{0.5em} } }

\newcommand{\squishend}{\end{list}}

\newcommand{\PP}[1]{
	\noindent{\bf{#1}.}\xspace
}

\SetCommentSty{mycommfont}

\newcommand{\thickhline}{%
	\noalign {\ifnum 0=`}\fi \hline height 1pt
	\futurelet \reserved@a \@xhline
}

\makeatletter
\newcommand{\linebreakand}{%
\end{@IEEEauthorhalign}
\hfill\mbox{}\par
\mbox{}\hfill\begin{@IEEEauthorhalign}
}
\makeatother

\begin{document}

\title{\paperTitle}

\author{\IEEEauthorblockN{Abhijit Suprem}
\IEEEauthorblockA{\textit{School of Computer Science} \\
\textit{Georgia Institute of Technology}\\
Atlanta, USA\\
asuprem@gatech.edu}
\and
\IEEEauthorblockN{Rodrigo Alves Lima}
\IEEEauthorblockA{\textit{School of Computer Science} \\
	\textit{Georgia Institute of Technology}\\
	Atlanta, USA\\
	ral@gatech.edu}
\and
\IEEEauthorblockN{Bruno Padilha}
\IEEEauthorblockA{\textit{Department of Computer Science} \\
	\textit{University of São Paulo}\\
	São Paulo, Brazil\\
	brunopadilha@usp.br}
 \linebreakand 

\IEEEauthorblockN{João Eduardo Ferreira}
\IEEEauthorblockA{\textit{Institute of Mathematics and Statistics} \\
	\textit{University of São Paulo}\\
	São Paulo, Brazil\\
	jef@ime.usp.br}
\and
\IEEEauthorblockN{Calton Pu}
\IEEEauthorblockA{\textit{School of Computer Science} \\
\textit{Georgia Institute of Technology}\\
Atlanta, USA\\
calton.pu@cc.gatech.edu}
}

\maketitle

\begin{abstract}
	As camera networks have become more ubiquitous over the past decade, the research interest in video management has shifted to analytics on multi-camera networks. This includes performing tasks such as object detection, attribute identification, and vehicle/person tracking across different cameras without overlap. Current frameworks for management are designed for multi-camera networks in a closed dataset environment where there is limited variability in cameras and characteristics of the surveillance environment are well known. Furthermore, current frameworks are designed for offline analytics with guidance from human operators for forensic applications. This paper presents a teamed classifier framework for video analytics in heterogeneous many-camera networks with adversarial conditions such as multi-scale, multi-resolution cameras capturing the environment with varying occlusion, blur, and orientations. We describe an implementation for vehicle tracking and vehicle re-identification (re-id), where we implement a zero-shot learning (ZSL) system that performs automated tracking of all vehicles all the time. Our evaluations on VeRi-776 and Cars196 show the teamed classifier framework is robust to adversarial conditions, extensible to changing video characteristics such as new vehicle types/brands and new cameras, and offers real-time performance compared to current offline video analytics approaches.

\end{abstract}


\section{Introduction}
The ubiquity of large-scale camera networks have coincided with the emergence of real-time video management platforms such as Chameleon \cite{RN49}, Kestrel \cite{RN50}, VideoStorm \cite{RN42}, and BriefCam \cite{RN51}. These tools allow human users to manage camera networks of thousands to hundreds of thousands of cameras, and query them manually to obtain live and archived video data summarization, mainly for forensics applications. 

From the big data and machine learning (ML) research point of view, a major research challenge is the automation of video analytics to detect and track interesting objects and events. Metadata extraction and object tracking must be robust to streams with varying resolution and scale, along with  camera artifacts such as different levels of blur and orientations. For practical applications, ML models also need to be extensible, so more classes of objects can be added as they become interesting. Finally, for real-time analytics, such tracking requires fast models due to the amount of data to be analyzed within a limited time (milliseconds per frame). 

A typical vehicle tracking system consists of: (i) vehicle detection from frames, (ii) collection and integration of detections and metadata on the identified vehicle, and (iii) vehicle re-identification across different frames and cameras. Various approaches have been proposed (we cover Related Work in \cref{sec:related}), but they have performance issues due to significant computation requirements and extensibility issues due to assumptions on the vehicles, cameras, or other system components. To automate the process, effective knowledge acquisition models are necessary to detect and track relevant objects, infer missing metadata, and enable automated event detection.

In this paper, we propose reframing the typical large-scale video analytics pipeline from the common single-model approach to a novel teamed-classifier approach to deal with real-world video datasets with dynamic distributions. In a teamed-classifier approach, we build teams of models where each model, or a subset of models, is assigned to a subspace in the data distribution. We can contrast with ensembles. In a traditional bagging or stacked ensemble, each model in the ensemble is applied to the entire data space and weighted on either training performance or live drift-detected performance (dynamic weighted ensembles for drifting data are presented in \cite{RN66, RN65}). The prediction for an input $x$ is then: $\hat{y}=\sum_{i=1}^{m} \alpha_i h_i(x)$, where $\alpha_i$ is a static weight for model $h_i\in\{h_1,h_2,\cdots,h_m\}$, with $\alpha_i$ usually assigned empirically. In contrast, our teamed classifier approach assigns an expert, or a family of expert classifiers, to a region of the input data space. We use a gating function to dynamically construct an ensemble during inference, similar to \cite{RN61}:

\begin{equation}
\label{eq:teamed}
\hat{y}=\sum_{i=1}^{m} g(x)h_i(x)
\end{equation}

Our contributions can be summarized as follows:

\squishlist
\item A teamed classifier approach for video analytics, with focus on the vehicle re-identification task. 
\item A general framework for re-identification that uses the naturally induces sparsity of the re-id task to build a sparse gating model with supervision. We evaluate our gating model on the Cars196 Zero-Shot learning task, where the goal is to cluster vehicle brands and models. We achieve state-of-the-art performance with normalized mutual information (NMI, used to evaluate clustering quality) of 66.03 compared to previous SoTA of NMI 64.4 \cite{RN72} and NMI 64.90 \cite{RN70}).
\item A simple and strong baseline algorithm for the re-id task that can operate on the subspaces identified by the gating function. Our simple and strong base model is competitive with current state-of-the-art with an order of magnitude fewer parameters: we use approximately 12M parameters for our base model to achieve 64.4 mAP compared with more than 100M parameters for MTML-OSG \cite{RN19} with 62.6 mAP (mAP, or mean average precision, is a metric for evaluating ranking and retrieval).
\squishend

The teamed classifier approach addresses two intertwined technical challenges of high inter-class similarity and high intra-class variability:
\squishitemize
\item \textbf{High inter-class similarity}: The visual similarity of two different vehicle of the same model/year and color, due to the manufacturing process of vehicles. Therefore, identifying the vehicle model/year and color (the best a pixel-based image analysis algorithm can do) would still occasionally be insufficient when two such vehicles appear in the same frame. 
\item \textbf{High intra-class variability}: Images of the same vehicle may look very different due to different orientations or environmental occlusion. For example, front-view image of a sedan looks quite different from rear-view image of the same sedan. Pure pixel-based image analysis may have difficulties with such differences. 
\squishend

\section{Related Work}
\label{sec:related}
\subsection{Classic Vehicle Tracking Approaches and Research Issues}
\label{sec:relatedtracking}
A typical vehicle tracking system consists of several stages:
\squishitemize
\item \textbf{Object detection}: Dense object detection in video streams has long been an integral part of video analytics. Various approaches have been devised for real-time detection of a wide variety of object types, such as vehicles, people, animals, and traffic signs, such as Mask-RCNN \cite{RN52}, YOLO\cite{RN30}, SSD-on-MobileNets \cite{RN54}.

\item \textbf{Vehicle metadata}: A important part of traffic management is tracking vehicle speed to ensure traffic safety and detecting speed limit infractions. While some specialized cameras in a surveillance network may be equipped with speed radar, such functionality is usually not found in common surveillance camera models. Recently, some proposed approaches perform speed detection in common cameras by  tracking the vehicles in 3D space \cite{RN32}. Other metadata include vehicle type, brand, and color.

\item \textbf{Vehicle re-identification}: The re-identification task requires tracking vehicles across cameras and assigning them to correct identities. Challenges lie in re-id under adversarial conditions where vehicles need to be tracked with multi-orientation, multi-scale, multi-resolution values alongside possible occlusion and blur. The vehicle re-identification problem has seen significant work in the past few years due to advances in the general one-shot learning problem \cite{RN19, RN27, RN2, RN8, RN9, RN3, RN4}. 

\item \textbf{Event detection}: Automated event detection remains a difficult challenge due to the lack of labeled real-world or synthetic data and absence of frameworks for video-based anomaly detection. A few approaches have been tested on simpler, small-scale data, such as LSTM-based \cite{RN57} or predictive coding \cite{RN55} approach.
\squishend

There have been several advancements towards some of these tasks. The video-management platforms mentioned also perform object detection using off-the-shelf models. For example, \cite{RN51} uses pretrained detection models and performs time-stamp base summarization, similar to pretrained YOLO for vehicle detection in \cite{RN50}. The approach in \cite{RN49} uses both YOLO and Mask-RCNN for object detection \cite{RN49}.

\begin{figure*}[t]
	\centering
	\includegraphics[width=5.0in]{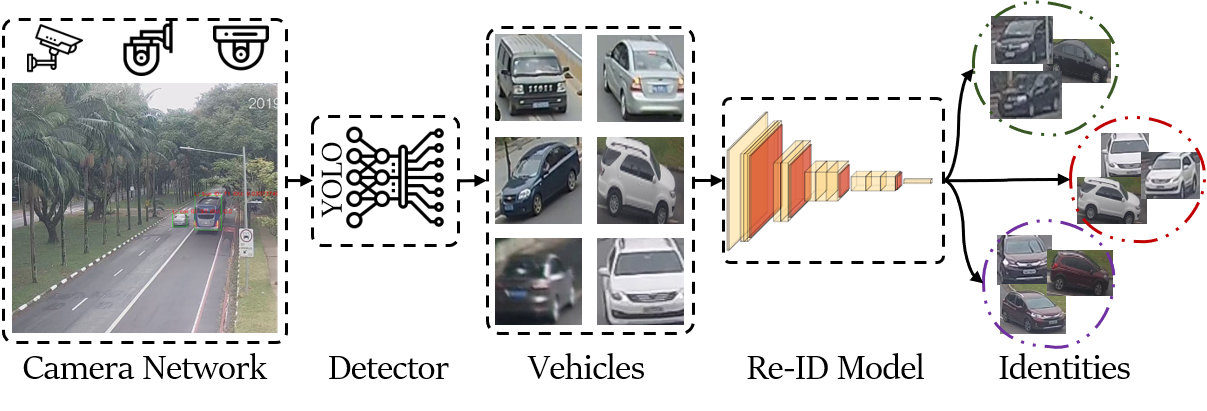}
	\caption{The typical vehicle re-identification pipeline. Each layer extractors progressively finer features.}
	\label{fig:typicalpipeline}
\end{figure*}

\PP{Vehicle re-identification} More recently, there have been approaches for end-to-end vehicle metadata extraction and re-identification in \cite{RN19, RN2, RN8, RN7, RN3, RN4}. OIFE \cite{RN7} proposed stacked convolutional networks (SCN) to extract fine-grained features in conjunction with global features. 20 such keypoints on vehicles, such as headlights, mirrors, and emblem were labeled and extracted by SCNs to build feature masks. Global features and masked features are combined to create orientation invariant features for re-id. RAM \cite{RN2} approaches fine-grained feature extraction by splitting vehicle images into three regions and extracting features from each region separately. Features are combined with a fully-connected network for re-id. VAMI \cite{RN3} adds additional supervision to fine-grained feature extraction by using the viewpoint information of vehicles. Subnetworks are built for each vehicle viewpoint, and features from view-point subnetworks are combined for re-id. EALN \cite{RN8} proposes addressing inter-class similarity with intra-class variability by using generated negatives: by using GANs to reconstruct images of existing vehicles, EALN can create potentially infinite training samples from a small dataset to improve inter-class similarity discrimination. MTML \cite{RN19} combines ideas from RAM and VAMI by creating subnetworks for different orientations, scales, and color corrections. Features from subnetworks are combined for re-id. Finally, QD-DFL \cite{RN4} proposes retaining spatial information in features by extracting diagonal and anti-diagonal features. Instead of only flattening convolutional features, diagonal features values are also used to improve re-id.

\PP{Single Models in Video Analytics} A common theme in the typical methods is the use of a single network for each task (\cref{fig:typicalpipeline}). Kestrel \cite{RN50} uses the same model for all vehicle tracking. While Chameleon uses differently-sized models for low-fps, medium-fps, and high-fps streams, there is little variability beyond this – a single Mask-RCNN model is used for all high-fps videos, for example. Finally, most re-identification models propose a single network for all types of vehicles to perform simultaneous vehicle attribute extraction and identification. Each of the re-id models discussed uses end-to-end training; while a model may have subnetworks, they are used as a single model for each sample. Such approaches are effective in small-scale datasets without much variability. However, large-scale, real-world video networks have a variety of adversarial conditions that can limit single-model effectiveness. A demonstration of such adversarial condition based model degradation is provided in \cite{RN37}, where the authors examine some state-of-the-art re-identification models on adversarial re-id datasets with multi-scale, multi-resolution images along with occlusion and motion blur and find performance deterioration due to high dataset variability. Similarly, recent research in domain adaptation \cite{RN23, RN25} show different datasets that are visually similar for humans encode artifacts that can cause significant model deterioration. The authors of BlazeIt \cite{RN58}, a video querying framework, also make such an observation: video drift due to changes in the data distribution can lead to model performance degradation unless new models are added.

\PP{Open and Closed Datasets}
One of the important issues in automated video analytics is the distinction between closed datasets that have finite underlying features under non-adversarial conditions and open datasets that are continuously evolving with potentially infinite underlying features. Most datasets used to train vehicle re-id or object detection models are closed datasets: their class distributions are fixed, and they encode a static set of features. This can lead to development of models that do not generalize to real-world data. Findings in generalizability studies in \cite{RN59} and \cite{RN60} show model iteration on closed datasets lead to architectures and model weights that perform well on their respective test sets without generalizing to real-world data in the same domain (CIFAR-10 and ImageNet, respectively). This supports the findings in \cite{RN23} where models trained on one person re-id dataset significantly underperform on another, visually indistinguishable person re-id dataset. It is evident, then, that real-world analytics must take into account the open nature of real-world data where the underlying feature distribution is continuously evolving \cite{RN61} and dataset drift is commonplace \cite{RN62, RN58, RN64}.

\subsection{Research Issues in Teamed Classifiers}
\label{sec:team}

\begin{figure*}[t]
	\centering
	\includegraphics[width=6.5in]{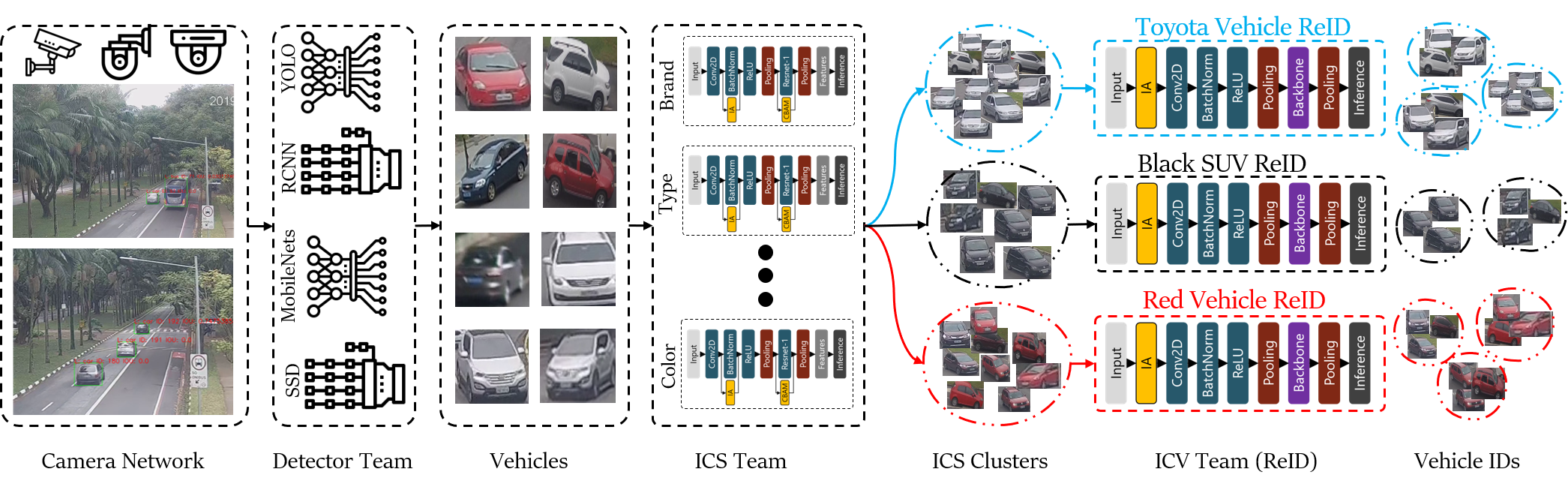}
	\caption{Teamed Classifiers for Vehicle Identification System; ICS: Inter-Class Similarity; ICV: Intra-Class Variability.}
	\label{fig:teamedpipeline}
\end{figure*}

\PP{Team Sparsity}
An important consideration in teamed classifiers is sparsity of weight assignments from the gating function $g(x)$ from \cref{eq:teamed} to ensure any single sample x uses only a subset of models. Differently from the recently proposed sparse mixture-of-experts model \cite{RN67}, subspace model assignments in teamed classifiers are supervised. In the mixture-of-experts approach, the submodels and gating function for submodels are trained together and sparsity is enforced with a penalty term in the loss function. In our teamed classifier approach, we enforce sparsity by exploiting \textit{naturally induced sparsity} in our input space; for example, vehicle re-identification has naturally induced sparsity in the manufacturing process: a vehicle must be of a single type (sedan or SUV), and of a single brand (Toyota, Mazda). So, we construct a supervised gating function that ensures sparsity using this natural sparsity in the input space by detecting vehicle brand, then using a brand-specific expert. We again contrast with the sparse mixture-of-experts model, where gating functions and experts and trained together to let the gating function learn the subspace assignments without supervision \cite{RN67, RN68, RN69}. Adding new subspaces or changing existing subspaces, as is the case with real-world drift \cite{RN61, RN64}, requires retraining the entire mixture-of-experts. In the teamed classifier approach, we can train the gating function and experts independently, allowing us to more easily extend to new subspaces by creating new experts as and when required and training them independently of existing experts. Changes to an existing subspace require only updating that subspace's assigned models.

\PP{Naturally Induced Sparsity}
We consider the naturally induced sparsity in vehicle tracking. The vehicle tracking task requires clustering vehicle identities into disjoint groups such that all images of a single identity are identified as such. This research task involves two technical challenges: high inter-class similarity (two vehicles of the same model/year and color are visually the same by manufacturing process), and high intra-class variability (images of the same vehicle from different perspectives can look very different). 

The naturally induced sparsity of the re-id task lies in high inter-class similarity, since we observe that the inter-class similarity problem is precisely due to the underlying manufacturing process; some examples of inter-class similarity clusters include groups of Toyota Corollas, black SUVs, or red vehicles. Conversely, existing vehicle re-id datasets such as VeRi-776 \cite{RN1} and VeRi-Wild \cite{RN38} primarily focus on intra-class variability. Current approaches in vehicle re-id attempt to address inter-class similarity and intra-class variability in the same end-to-end model \cite{RN2, RN8, RN9, RN4}. This creates models that sacrifice performance on solving edge cases in intra-class variability to increase discriminative ability for inter-class similarity across the entire data space.

\section{The Teamed Classifier Approach}
\label{sec:teamedclassifier}

We introduce our teamed classifier approach for video analytics, specifically for vehicle re-identification. We will first describe a typical video analytics pipeline for re-identification. Then we describe our teamed classifier approach for vehicle re-identification and the advantages it brings over traditional single-model pipelines.

\subsection{Typical Pipeline for Vehicle Re-ID}
\label{sec:typicalpipeline}

A typical video surveillance framework for re-id comprises of a pipeline of increasingly fine-grained feature extractors. We show in \cref{fig:typicalpipeline} a standard vehicle surveillance pipeline. Data enters the pipeline through a deployed camera network in the form of video streams. Each layer  performs progressively finer-grained feature extraction for knowledge acquisition. 

In the Object Detector layer, pretrained object detectors such as YOLO \cite{RN30} or Mask-RCNN \cite{RN52} are commonly used for vehicle, person, and sign detection. As we have discussed, the usual approach in current systems such as Kestrel and VideoStorm is to use a single model type for the entire data space. A notable exception is Chameleon \cite{RN49}, which uses a small team of detectors for changes in detection quality requirements: if high-quality detections are requested, a pretrained YOLO detector is used. If low-quality detections are requested, then simpler detectors like SIFT are used. Object detectors extract very coarse features, namely labels.

The Re-ID Model layer is the focus of typical re-id approaches, where a single end-to-end model is developed for fine-grained vehicle identity clustering. Details about these end-to-end models are provided in Related Work. Here we observe that some approaches do use submodels, such as OIFE \cite{RN7}; however these submodels are trained together and are each designed for the entire input space. Each submodel’s features are subsequently combined with an additional dense neural network to obtain final re-identification features. Re-id features are clustered to identify unique vehicle identities.

\subsection{Teamed Classifiers for Vehicle Re-ID}
\label{sec:teamedreid}
We now present our teamed classifier-based pipeline for a vehicle identification system in \cref{fig:teamedpipeline}. Our approach differs from the current methods described in \cref{fig:typicalpipeline} by employing classifier teams as feature extractors, where each member of the team is assigned to a different region of the input space. We exploit the naturally induced sparsity of the input space to create disjoint teams with supervision.

\PP{Detector Team}
We employ detector teams in lieu of cross-domain adaptive object detector models. A challenge in real-world object detection on multi-stream video networks is the sharp difference between frame artifacts generated by each camera or set of cameras. As the analysis of cross-domain performance in \cite{RN24} shows, even visually similar images are difficult for feature extractors if they are captured in different environments. While there have been some studies in developing domain-adaptive techniques or more generalizable universal detector models \cite{RN73}, we employ the student-teacher model for object detection from \cite{RN74}. We use a pretrained full YOLOv3 model as the teacher, and train smaller, specialized detectors for each camera. The specialized models are built on SSD-MobileNets and can be deployed on embedded devices \cite{RN76, RN54}. Specialized detectors are covered in recent approaches; we focus on the identification layers.

\begin{figure}[t]
	\centering
	\includegraphics[width=2.4in]{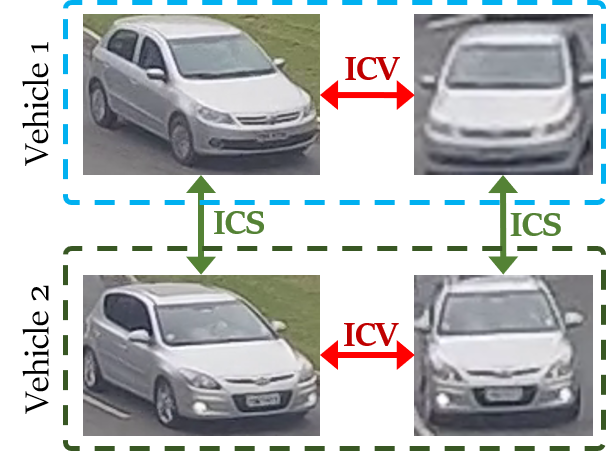}
	\caption{\textbf{Inter-class similarity \& intra-class variability}. Vehicle 1 is visually similar to Vehicle 2 and can be differentiated only by looking at windows and bumper. This is an example of inter-class similarity in white SUVs}
	\label{fig:icsicv}
\end{figure}

\PP{Inter-Class Similarity Team}
For vehicle re-id significant interest has been given towards developing models that can handle both inter-class similarity and intra-class variability, shown in \cref{fig:icsicv}. In the former, vehicles with different identities (i.e. license plates) look very similar because they may be from the same brand, same vehicle type, or same color. A Re-ID model must therefore distinguish visually similar vehicles in the same camera using camera-specific artifacts such as spatio-temporal constraints or background information, while also capturing cross-camera features. In terms of implementation, a Re-ID model must generate a set of features for vehicles such that similar vehicles across multiple orientations, resolutions, and scales are projected to the same cluster, while ensuring features of different vehicles that have high inter-class similarity are projected to different clusters.

This naturally imposes orthogonal constraints on a re-id model, and fine-grained feature extraction is necessary to ensure a model can address both constraints. Thus, a model must be able to capture the full range of feature combinations in vehicles across multiple brands, orientations, colors, resolutions, and scales, as have been proposed in existing approaches. Consequently, existing approaches build complex networks that perform inter-class similarity discrimination, and intra-class variability minimization in the same model: OIFE \cite{RN7} uses 20 stacked convolutional networks to extract human-labeled keypoints; RAM \cite{RN2} builds three sub-networks to evaluate each section of a vehicle (roof, body, chassis); VAMI \cite{RN3} creates multiple sub-models for each orientation; and QD-DLF \cite{RN4} builds four  networks to extract diagonal  features.

\begin{figure}[h]
	\centering
	\includegraphics[width=3.2in]{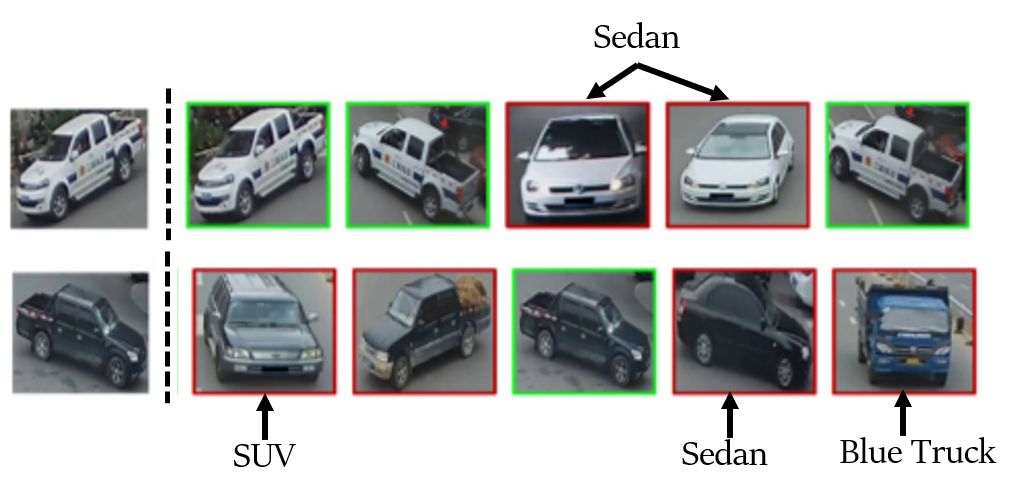}
	\caption{Mistakes in state-of-the-art Group Sensitive Triplet Learning \cite{RN20}; images taken directly from original paper. The first row shows retrievals of sedans for a query of truck. The second row shows retrievals or SUV, sedan, and a blue truck for query of black truck.}
	\label{fig:simplemistakes}
\end{figure}

While such approaches partially address the inter-class similarity and intra-class variability constraints, they make simple mistakes: we show in \cref{fig:simplemistakes} some mistakes in vehicle re-id provided by the Group Sensitive Triplet Embedding approach in \cite{RN20}. Similar examples are provided in other papers. We observe that forcing a re-id model to learn both inter-class similarity discrimination and intra-class variability minimization enforces a learning burden that reduces overall performance.

We again propose exploiting the naturally induced sparsity of the input space to reduce the burden of learning orthogonal feature extraction for the re-id model. Concretely, our teamed-classifier approach uses two layers of features extractors: an inter-class similarity team to perform coarse clustering of vehicle images using natural feature descriptions, followed by an intra-class variability team  that assigns one re-id model to each cluster from the inter-class similarity team. This allows the re-id models in the intra-class variability team to focus on a subset of the input space of vehicles without enforcing a generalization constraint to address inter-class similarity. 

\PP{Naturally Induced Sparsity}
We use our observations from \cref{fig:simplemistakes} to build the inter-class similarity team; we select three key coarse features for enforcing the intra-class variability team sparsity – vehicle color, vehicle type, and vehicle model. We focus on vehicle model discrimination, since vehicle color and type are coarser, finite features addressed with simpler image classifiers as in the BoxCars116K models \cite{RN77}. For vehicle model discrimination, we consider the related zero-shot learning task. The zero-shot learning task requires learning feature extractors that can discriminate between classes seen during training and generalize to unseen classes not seen during training. We specifically focus on the Cars196 dataset, since it requires identifying unseen vehicle models using feature extraction on seen vehicle models. This is useful in re-id since new vehicle brands and updated vehicle models are continuously introduced, adding dataset drift to the input space. 

We develop a zero-shot learning model that achieves state-of-the-art performance on the Cars196 dataset and use it for model discrimination. Our model implicitly learns relevant features for the unsupervised clustering of vehicle models. We describe our vehicle brand discriminator in \cref{sec:icsmodel}.

\PP{Intra-Class Variability  Team}
With an inter-class similarity team to perform coarse-grained clustering, we can build our re-id models to focus on minimizing intra-class variability only. This provides two advantages:

\squishlist
\item Since our models only need to address intra-class variability on a limited subset of the true input space, we achieve higher performance in mAP and rank-1 retrieval compared to recent approaches.
\item We can build smaller models compared to recent approaches. As such, each member of the intra-class variability team uses a single ResNet 18 backbone and can operate in near real-time, compared to 20 stacked convolutional networks in \cite{RN7}, 5 ResNet backbones in \cite{RN2}, and 4 ResNet50 backbones in \cite{RN4}.
\squishend

We describe our intra-class variability team's base model in \cref{sec:icvmodel}.
\section{Intra-Class Similarity Team}
\label{sec:icsteam}
We develop an end-to-end model to deploy as a submodel in the intra-class similarity team using a single backbone network. Our approach implicitly learns relevant local and global features for unsupervised clustering without relying on data and feature augmentation or synthetic data. We first describe the Cars196 dataset we use for evaluating our intra-class similarity team’s brand discrimination models.

\subsection{Dataset and Evaluation}
\label{sec:icsdata}
The Cars196 dataset, introduced in \cite{RN78}, contains 196 classes of vehicles. It is challenging due to few images per class (on average, Cars196 has 82 images per class). Furthermore, vehicles exhibit a high degree of inter-class similarity as described in \cref{sec:teamedreid}, since most vehicles fit into a few form factors. We evaluate our models with two metrics: the normalized mutual information measure and the top-1 retrieval rate (we also show results for top-5 retrieval).

\PP{Normalized Mutual Information (NMI)}
NMI measures clustering correlation between a predicted cluster set and ground truth cluster set; it is the ratio of mutual clustering information and the ground truth, and their harmonic mean. Given a set of predicted clusters $\Omega=\{c_1,c_2,\cdots,c_k\}$ under a $k$-means clustering, we say that each $c_1$ contains instances determined to be of the same class. With ground truth clusters $\mathds{C}=\{c'_1\,c'_2,\cdots,c'_k\}$, we calculate NMI as:

\begin{equation}
\label{eq:nmi}
NMI(\Omega, \mathds{C})=\frac{2I\Omega,\mathds{C}}{H(\Omega)+H(\mathds{C})}
\end{equation}

where $H(\cdot)$ is the entropy and $I(\cdot)$ is the mutual information between $\Omega$ and $\mathds{C}$. Since NMI is invariant to label index, no alignment is necessary.

\PP{Top-k Retrieval}
We use the standard top-$k$ ranking retrieval accuracy for, calculated as the percentage of classes correctly retrieved at the first rank, and of those missed, percentage retrieved correctly on the second rank, and so on.

\subsection{Model for Brand Discrimination}
\label{sec:icsmodel}
We make the following observations in creating our intra-class similarity team’s submodel for brand discrimination:

\squishlist
\item It is well known that the earlier kernels in a convolutional network learn abstract, simple features such as colors. Some kernels also learn basic geometric shapes corresponding to image features in the low frequency range of images.
\item Later kernels learn more class-specific details and extract detailed features corresponding to higher-frequency features. While these are useful for traditional image classification, they may overfit on the ZSL task.
\item Convolutional layers are used for feature \textit{extraction}, and subsequent dense, or fully-connected layers, used for feature \textit{interpretation}. This forces the dense layers to learn image feature discrimination, instead of relying on convolutional filters. Since convolutional filters focus on nearby pixels with a spatial constraint, we believe relying on convolutional filters for feature interpretation to be more effective in tracking image invariant features.
\squishend

Since the earlier layers learn coarse features, we propose using the early kernels with attention modules from \cite{RN11} to improve feature extraction. Since the later layers in a convolutional network already learn fine-grained features, they do not require augmentation; otherwise they would begin to overfit on the training data and fail to generalize to unseen brands. Thus, we use convolutional attention on the early layers only, as opposed to attention throughout. We also address the loss of spatial image features in dense layers by removing them entirely and only use convolutional layers for both feature extraction and interpretation: given query and target, we evaluate their similarity on only convolutional features. In contrast to current approaches that use dense layers after the convolutional backbone to perform feature interpretation, we force the convolutional network to also learn feature interpretation simultaneously with feature extraction. 

\begin{figure}[h]
	\centering
	\includegraphics[width=3.2in]{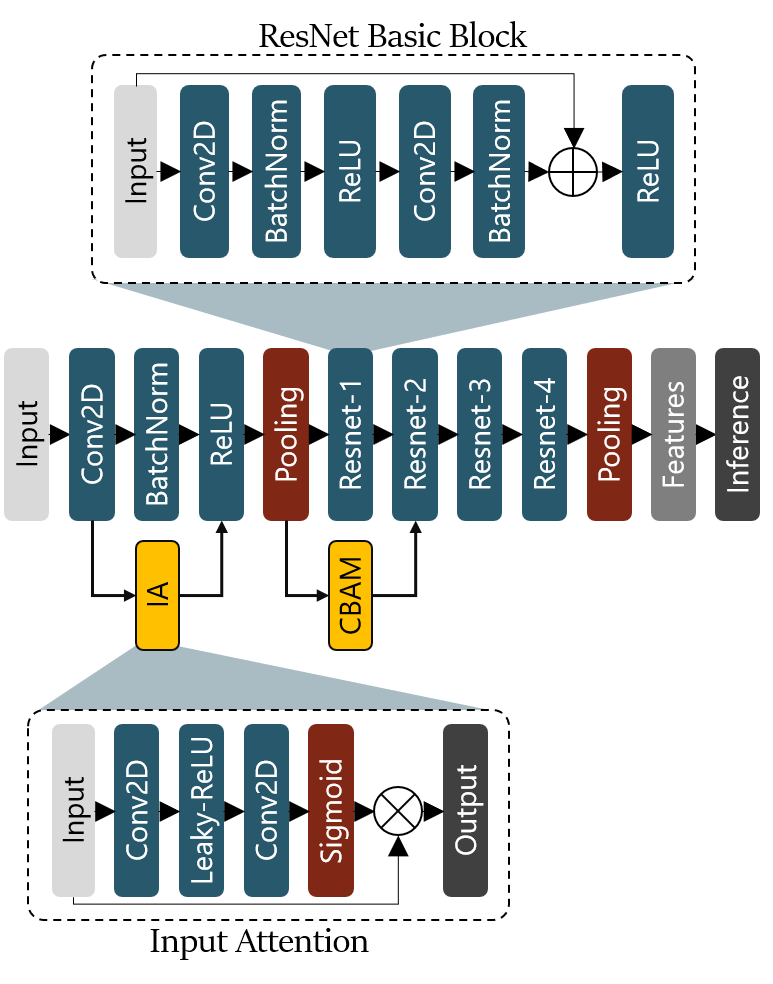}
	\caption{Overall architecture of intra-class similarity model for brand discrimination}
	\label{fig:icsmodel}
\end{figure}

We show our overall intra-class similarity model in \cref{fig:icsmodel}. We now describe the backbone and attention modules (CBAM \cite{RN11} and our novel Global Attention module).

\PP{Backbone}
We use ResNet-18 as the backbone for the intra-class similarity model. Each ResNet-18 backbone consists of several "basic blocks" chained together. We apply targeted attention to these basic blocks, as opposed to each convolutional layer.

\PP{Convolutional Attention}
We use the CBAM attention module from \cite{RN11} to add discriminative ability to the backbone. Since the earlier filters learn coarser features and the later filters learn fine-grained features, adding convolutional attention to all layers improves classification accuracy in general. However, in the brand discrimination task, we need to generalize to unseen brands; so CBAM on later layers causes networks to overfit on fine-grained features of the training set, reducing overall performance (we examine this performance drop in \cref{sec:icvevaluation}). We add attention only to the first basic block to learn discriminative coarse-grained filters for better unseen class separation and metric learning, improving performance in for Cars196 and brand discrimination in general.

\PP{Global Attention}
The CBAM block is not sufficient to improve generalization due to skewed feature maps in early convolutional layers. The first convolutional layer is crucial in feature extraction since it occurs at the beginning of the network. We find that many feature maps at the first layers do not track any useful features; instead they either output random noise or focus on irrelevant features such as shadow. Therefore, we develop the Global Attention module (GA), shown in to perform feature map regularization. The GA module reduced feature map skew by re-weighting feature weights. Whereas CBAM separates channel and spatial attention, GA combines them to ensure spatial features are learned together. GA uses two $3\times3$ conv layers with Leaky ReLU activation to retain negative weights from the first conv layer in the backbone. The output is passed through a sigmoid activation and element-wise multiplied with the  (see GA inset in \cref{fig:icsmodel}).

\begin{figure}[h]
	\centering
	\includegraphics[width=3.2in]{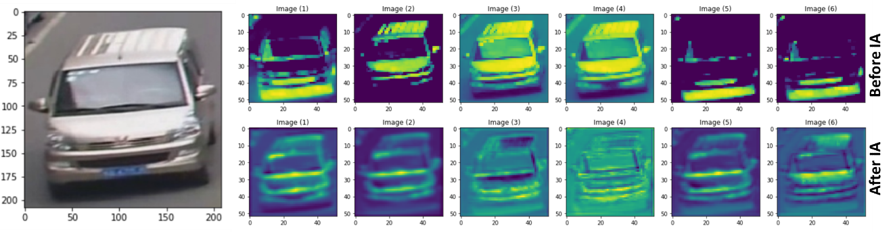}
	\caption{Feature map correction after Global Attentio (GA)n. Top features are skewed towards dark shadow, which does not carry any information. GA drives attention towards more useful details and corrects the skew.}
	\label{fig:iacorrection}
\end{figure}

We avoid max and average pooling since they cause loss of information and we want to preserve discriminative features for the basic blocks in the architecture core. We show an example of feature map correction in \cref{fig:iacorrection}; the top layer shows original feature maps, which have skewed towards the shadow under the vehicle; the bottom layer shows the corrected feature map without skew after applying GA.

\PP{ProxyNCA Loss}
Our brand discrimination model’s target is to learn a distance metric on vehicle brands such that features for vehicle brands are clustered in the output feature space. We train with the ProxyNCA loss for distance metric learning introduced in \cite{RN70}. With ProxyNCA, a model learns a set of proxies that map to training classes. Features from training-set images are mapped to the proxy-domain and the distance is maximized amongst the proxy clusters themselves. During inference, the proxies are ignored, and the true features are used to cluster vehicle brands.

\section{Intra-Class Variability Team}
\label{sec:icvteam}
We now describe our base model re-id for the intra-class variability team. First, we describe the VeRi-776 dataset we use for evaluating our model. Then we will cover our intra-class variability results.

\subsection{VeRi-776 Dataset}
\label{sec:veri}
The VeRi-776 dataset for vehicle re-id was introduced in \cite{RN1} to promote research in the field. It contains 776 unique vehicle identities, with 576 used for training and the remaining 200 used for testing. During testing, the target is to retrieve the unseen identities given a query, with performance evaluated on the ranking. The dataset contains mostly intra-class variability, with each identity having several images from multiple cameras in different environmental conditions. 

\subsection{Base Model for Re-ID}
\label{sec:icvmodel}
We now describe our base re-id model for the intra-class variability team. Since we offload inter-class similarity discrimination to the inter-class similarity team, our  re-id models are simpler and smaller than typical re-id models, with better performance. The intra-class variability base model is \textbf{r}obust, \textbf{e}xtensible, and \textbf{f}ast, as we will show. We call it REF-GLAMOR, for reference-GLAMOR, where GLAMOR stands for \textbf{Gl}obal \textbf{A}ttention \textbf{Mo}dule for \textbf{R}e-ID.

\PP{Base Model Construction}
We follow similar principles in designing the  re-id models that we used in the inter-class similarity brand discrimination model. Specifically, we use the ResNet-18 backbone with GA. Differently from the inter-class similarity models:
\squishlist
\item We do not use CBAM on the   re-id models. Since CBAM has already been applied to the first basic block in inter-class similarity models, CBAM on the first basic block in a  re-id model would perform redundant feature extraction. CBAM on the last basic block would lead to re-id model overfit on the training data. 
\item We use the warmup learning rate suggested in \cite{RN43} to gradually increase the learning rate during training and improve feature extraction on pretrained backbones.
\item We use the Random Erasing Augmentation proposed in \cite{RN15}. Surprisingly, we found this method, which has been shown to be effective in face recognition, has not been applied to vehicle re-id.
\squishend

\PP{Triplet Loss}
We use the standard triplet loss for distance metric learning on the features:

\begin{equation}
\label{eq:triplet}
L_T=\sum ||f(a)-f(p)||_2^2 - ||f(a)-f(n)||_2^2 + \alpha
\end{equation}

where $a$,$p$,$n$ are the anchor, positive, and negative of a triplet, $f(\star)$ is the embedding network, and $\alpha$ is the margin constraint enforcing the minimum distance difference between two images from the save identity (anchor and positive) compared to two images from distinct identities (anchor and negative). The triplet loss generates a mapping $f:\mathds{R}^{image}\rightarrow\mathds{R}^{embedding}$ such that individual identities are mapped to the same point.
\section{Evaluation}
To validate our approach, we evaluate each novel contribution: the inter-class similarity team for feature discrimination, and the intra-class variability team for re-id feature extraction. We compare our approaches to the state-of-the-art. 

\subsection{Brand Detection: Evaluation on Cars196}
\label{sec:icsevaluation}
Each submodel of the inter-class similarity team performance inter-class similarity clustering on natural features to enforce naturally induced sparsity on the subsequent intra-class variability team. As described in \cref{sec:icsdata}, we evaluate on the well-known Cars196 dataset with inter-class similarity.

\PP{Evaluation Results}
We evaluate NMI and Rank-1 on the Cars196 dataset and compare to recent approaches in the following table. We examine the impact of CBAM on different basic blocks to support our final model construction by testing different architectures: CBAM on all  blocks, CBAM on the first  block, and CBAM on the final  block of ResNet. 

Interestingly, addition of CBAM throughout the network reduces performance, since the later basic blocks begin overfitting on the fine-grained features that appear exclusively on the training set. We support this observation with results from CBAM-1, where CBAM is applied to the first basic block, and CBAM-4, where CBAM is applied to the final (or fourth) basic block. The results support our observations in \cref{sec:icsmodel} - CBAM-4 has even worse performance, while CBAM-1 increases performance beyond CBAM everywhere. 

On our model with CBAM-1 and global attention, we achieve state-of-the-art results (Table I), with NMI 66.03\% and Rank-1 of 82.75\% (10\% better than  \cite{RN70}).

\begin{table}[h]
	\centering
	\label{tab:carszsl}
	\caption{Experimental results compared to recent approaches on the Cars196 dataset}
	\begin{tabular}{llllll}
		\hline
		\multicolumn{1}{c}{\textbf{Method}}                                          & \multicolumn{1}{c}{\textbf{NMI}} & \textbf{R-1}   & \textbf{R-2}   & \textbf{R-4}   & \textbf{R-8}   \\ \hline
		DML \cite{dml}                                                                         & \multicolumn{1}{c}{56.70}        & 49.50          & -             & -             & -             \\
		DSP  \cite{dsp}                                                                        & \multicolumn{1}{r}{64.40}        & 72.90          & 81.60          & 88.80          & -             \\
		Proxy-NCA \cite{RN70}                                                                       & \multicolumn{1}{r}{64.90}        & 73.22          & 82.42          & 86.36          & 88.68          \\ \hline
		Baseline (B)                                                                 & \multicolumn{1}{r}{54.63}        & 72.28          & 81.63          & 88.52          & 93.27          \\
		B+CBAM                                                                       & 56.71                            & 73.55          & 82.09          & 87.99          & 92.26          \\
		B+CBAM-4                                                                    & 31.39                            & 22.94          & 32.72          & 43.56          & 56.08          \\
		B+CBAM-1                                                                    & 63.04                            & 80.56          & 87.86          & 92.52          & 95.73          \\ \hline
		\textbf{\begin{tabular}[c]{@{}l@{}}B +CBAM-1 +GA\end{tabular}} & \textbf{66.03}                   & \textbf{82.75} & \textbf{89.68} & \textbf{93.72} & \textbf{96.41} \\ \hline
	\end{tabular}
\end{table}

\subsection{Re-ID: Evaluation on VeRi-776}
\label{sec:icvevaluation}
We now show performance of REF-GLAMOR for the intra-class variability minimization. We evaluate our overall model described in  \cref{sec:icvmodel} on the VeRi-776 dataset and compare to recent approaches. To evaluate the impact of IA, we compare performance to the re-id base model without global attention, and with global attention in Table II.

\begin{table}[h]
	\centering
	\label{tab:veri776}
	\caption{Performance of REF-GLAMOR base model on VeRi-776 re-id dataset}
	\begin{tabular}{lrrr}
		\hline
		\textbf{Approach}                                                                      & \textbf{mAP}   & \textbf{CMC-1} & \textbf{CMC-5} \\ \hline
		\begin{tabular}[c]{@{}l@{}}DAVR \cite{RN9}\end{tabular}           & 26.35          & 62.21          & 73.66          \\
		OIFE+ST \cite{RN7}                                                    & 51.42          & 68.30          & 89.70          \\
		\begin{tabular}[c]{@{}l@{}}Hard-View-EALN  \cite{RN8}\end{tabular} & 57.44          & 84.39          & 94.05          \\
		\begin{tabular}[c]{@{}l@{}}PATH-LSTM  \cite{RN6}\end{tabular}      & 58.27          & 83.49          & 90.04          \\
		GSTRE \cite{RN20}                                                     & 59.47          & \textbf{96.24} & \textbf{98.97} \\
		VAMI+ST \cite{RN3}                                                    & 61.32          & 85.92          & {\ul 97.70}    \\
		RAM \cite{RN2}                                                        & 61.50          & 88.60          & 94.00          \\
		\begin{tabular}[c]{@{}l@{}}Quadruplet  \cite{RN4}\end{tabular}     & 61.83          & 88.50          & 94.46          \\
		MTML-OSG \cite{RN19}                                                  & 64.6           & {\ul 92.30}    & 94.2           \\ \hline
		\begin{tabular}[c]{@{}l@{}}REF-GLAMOR  (Ours)\end{tabular}                            & 64.48          & 63.90          & 86.20          \\
		REF-GLAMOR(Ours)                                                                   & \textbf{71.08} & {\ul 89.21}    & {\ul 95.47}    \\ \hline
	\end{tabular}
\end{table}

\begin{figure}[h]
	\centering
	\includegraphics[width=3.2in]{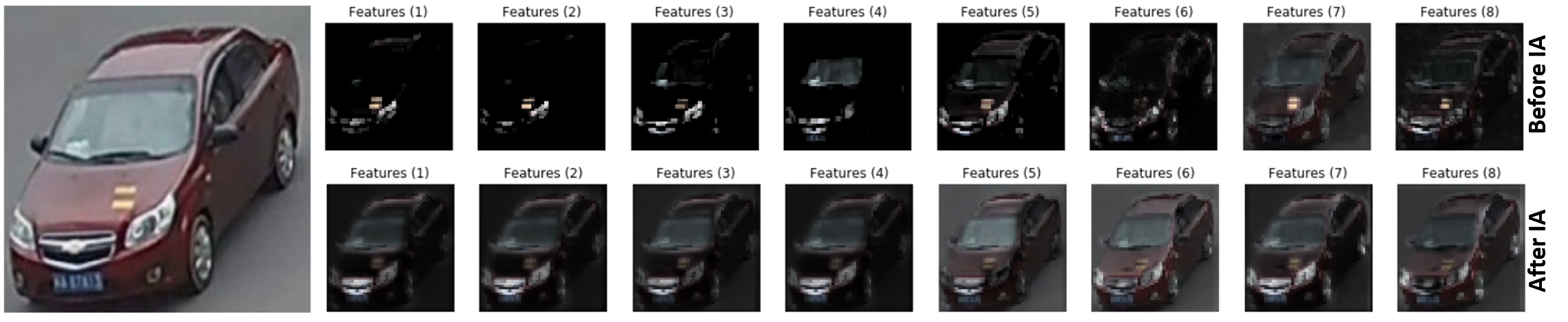}
	\caption{Feature map skew correction with Global Attention; loss of information due to poor filters is corrected in the first convolutional layer.}
	\label{fig:iacorrection2}
\end{figure}

REF-GLAMOR uses only a ResNet-18 backbone and achieves impressive performance compared to existing approaches that use multiple feature extractors. Since we do not need to perform inter-class similarity discrimination, REF-GLAMOR based models perform well on their respective subsets of the input space. Furthermore, performance is significantly improved with the inclusion of the GA module, since the first convolutional layer in the backbone has reduced feature map skew. We show an example in \cref{fig:iacorrection2}, where GA  reduces noisy or bad kernels in the first layer.

\section{Discussion}
\label{sec:discussion}
Here we discuss qualitative aspects of our teamed classifier approach for a vehicle ID framework. 

\subsection{Robustness of Re-ID Team}

\begin{table*}[t]
	\centering
	\label{tab:parameters}
	\caption{Approximate number of parameters in current re-id approaches compared to REF-GLAMOR}
	\begin{tabular}{lllrr}
		\hline
		\textbf{Approach}                     & \textbf{Construction}                                                                                                                                                                             & \textbf{Parameters}                                                                                                                                                  & \textbf{Total} & \textbf{mAP} \\ \hline
		OIFE \cite{RN7}      & \begin{tabular}[c]{@{}l@{}}20 stacked convolutional networks (SCN) for\\   feature maps, passed through 6 modified Inception \\   networks, with features combined with dense layers\end{tabular} & \begin{tabular}[c]{@{}l@{}}20x SCN: 26M params\\ 1x Modified Inception: $\sim$1M params\\ 1x Output FC: 1280x256 dense layer\end{tabular}                            & 521M           & 51.42        \\ \hline
		RAM \cite{RN2}       & \begin{tabular}[c]{@{}l@{}}3 ResNet50 branches, a normalization branch, and a global \\   features branch, with features combined with dense layers\end{tabular}                                  & \begin{tabular}[c]{@{}l@{}}3x ResNet50: 26M params\\ 1x Norm. branch: 80M params\\ 1x Features branch: 80M params\\ 1x Output FC: 5120x1024 dense layer\end{tabular} & 244M           & 61.5         \\ \hline
		MTML-OSG \cite{RN19} & \begin{tabular}[c]{@{}l@{}}4 convolutional feature extractors, each using ResNet50, \\   with features combined with dense (FC) layers\end{tabular}                                               & \begin{tabular}[c]{@{}l@{}}4x ResNet50: 26M params\\ 3x FC: 2048x575 dense layers\\ 1x Output FC: 8192x575 dense layer\end{tabular}                                  & 110M           & 64.6         \\ \hline
		REF-GLAMOR          & ResNet18 backbone + GA module                                                                                                                                                                     & \begin{tabular}[c]{@{}l@{}}1x ResNet18: 11M params\\ 1x GA Module: 200K params\end{tabular}                                                                          & 11M            & 71.08        \\ \hline
	\end{tabular}
\end{table*}

Our  team for vehicle re-id  is robust to multi-scale, multi-orientation images. This is evident from our results in \cref{tab:veri776}, where we show state-of-the-art mAP compared to existing approaches. While the CMC-1 is lower than a few methods like GSTRE \cite{RN20} and MTML-OSG \cite{RN19}, mAP is a better measure of robustness. Compared to top-k retrieval, which measures the recall at k-th ranking, mAP measures overall ranking quality by measuring fraction of true-positives in the retrieved results across all queries. Higher CMC-1 indicates the first result retrieved is relevant. Higher mAP indicates most top-ranked results retrieved are relevant. As such, it is a stronger measure of robustness. Our model with GA achieves overall robust performance at mAP 71.08, compared to the next best mAP of 64.6 from \cite{RN19}.

\subsection{Extensibility of Teamed Classifiers}
We have already discussed existing ensemble-based approaches (boosting/stacked ensembles) and the more recent mixture-of-experts ensembles. Our motivation in proposing teamed classifiers comes from our observation that several real-world domains have naturally induced sparsity.

Compared to the sparse mixture-of-experts model, which must learn the underlying sparse regions of the input space, our teamed classifier approach uses supervision to guarantee sparsity in the classifier teams. Furthermore, by separately training the gating function for brands/color and classifier teams for re-id, our pipeline is more extensible to new knowledge. Whereas the sparse mixture-of-experts must be retrained to handle new types of knowledge, our approach simply adds a new gate in the form of a new member to the intra-class similarity team, and corresponding classifiers for that gate to the re-id team. Further, as we demonstrated in the intra-class similarity evaluation, our individual team members handle unseen concepts.

Consequently, our re-id models in the re-id team need to only address intra-class variability. When inter-class similarity discrimination is performed by the intra-class similarity team, the gating models select which members of the re-id team will be used in re-ID. We show an example in the teamed classifier pipeline in \ref{fig:teamedpipeline}, where we depict three re-id teams (among many): the Toyota Brand team, the Red Vehicle Team, and the Black SUV Team. If a color discriminator model identifies a red vehicle, its respective re-id team is used to extract identification features. This allows us to reduce instances of re-id mistakes, as shown in \cref{fig:simplemistakes}.

\subsection{Real-time Performance}

By offloading inter-class similarity discrimination to the intra-class similarity team, we are able to make our re-id models smaller than existing approaches. We show in Table III the approximate number of parameters in current approaches and our own, along with overall mAP.

Since our re-id models use ResNet18, they can deliver real-time performance for vehicle tracking without GPUs, with reduced parameter ResNet18 \cite{RN80} achieving 50FPS on CPU. 
\section{Conclusion}
We have presented a new approach for conditional computation with teamed classifiers for vehicle tracking and identification. We describe an end-to-end approach for vehicle tracking, attribute extraction, and identification using teamed classifiers. With our teamed classifier approach, we build dynamic ensembles for feature extraction that are select at inference time. Similar to the mixture-of-experts model, we build a conditional network with sparsity that gates access to the dynamic ensembles. During pipeline construction, we build teams of models such that each member is assigned to a region of the input space. During inference, we determine the region of input space an input belongs to and dynamically select the team members for feature extraction. Unlike the mixture-of-experts model, where the sparsity constraint is enforced with training, our teamed classifier approach exploits the naturally induced sparsity of the input space in vehicle tracking to perform supervised team generation and gating.

We implement teamed classifiers for object detection (detector team models with camera-specialized detectors), vehicle attribute extraction, and vehicle identification. Since we adapt student teacher networks for the detector team and standard image classifiers for some attribute extractors, we focus evaluation on the novel team models: the brand discrimination team  and the re-id models. We demonstrate state-of-the-art performance on each task, and show the advantages of our teamed classifier approach in \cref{sec:discussion}, where we contrast the performance improvement of our approach with the reduced number of parameters (and consequently, operations), compared with current methods.

\section*{Acknowledgment}
This research has been partially funded by National Science Foundation by CISE’s SAVI/RCN (1402266, 1550379), CNS (1421561), CRISP (1541074), SaTC (1564097) programs, an REU supplement (1545173), and gifts, grants, or contracts from Fujitsu, HP, Intel, and Georgia Tech Foundation through the John P. Imlay, Jr. Chair endowment. Any opinions, findings, and conclusions or recommendations expressed in this material are those of the author(s) and do not necessarily reflect the views of the National Science Foundation or other funding agencies and companies mentioned above.

\bibliographystyle{IEEEtran}
\bibliography{cogmi}

\end{document}